\documentclass[conference]{IEEEtran}
\IEEEoverridecommandlockouts
\usepackage{cite}
\usepackage{amsmath,amssymb,amsfonts}
\usepackage{algorithmic}
\usepackage{graphicx}
\usepackage{textcomp}
\usepackage{xcolor}
\usepackage{csquotes}
\def\BibTeX{{\rm B\kern-.05em{\sc i\kern-.025em b}\kern-.08em
    T\kern-.1667em\lower.7ex\hbox{E}\kern-.125emX}}
\begin{document}

\title{Diverse Agents for Ad-Hoc Cooperation in Hanabi\\
}

\author{\IEEEauthorblockN{Rodrigo Canaan}
\IEEEauthorblockA{
\textit{NYU}\\
New York, USA \\
rodrigo.canaan@nyu.edu}
\\
\and
\IEEEauthorblockN{Julian Togelius}
\IEEEauthorblockA{
\textit{NYU}\\
New York, USA \\
julian.togelius@nyu.edu}
\\
\and
\IEEEauthorblockN{Andy Nealen}
\IEEEauthorblockA{
\textit{USC}\\
Los Angeles, USA \\
anealen@cinema.usc.edu}
\and
\IEEEauthorblockN{Stefan Menzel}
\IEEEauthorblockA{
\textit{HRI Europe GmbH}\\
Offenbach, Germany \\
stefan.menzel@honda-ri.de}
}

\IEEEpubid{\begin{minipage}{\textwidth}\ \\[12pt]
978-1-7281-1884-0/19/\$31.00 \copyright 2019 IEEE
\end{minipage}}
\maketitle

\begin{abstract}
In complex scenarios where a model of other actors is necessary to predict and interpret their actions, it is often desirable that the model works well with a wide variety of previously unknown actors. Hanabi is a card game that brings the problem of modeling other players to the forefront, but there is no agreement on how to best generate a pool of agents to use as partners in ad-hoc cooperation evaluation. This paper proposes Quality Diversity algorithms as a promising class of algorithms to generate populations for this purpose and shows an initial implementation of an agent generator based on this idea. We also discuss what metrics can be used to compare such generators, and how the proposed generator could be leveraged to help build adaptive agents for the game. 
\end{abstract}

\section{Introduction}

Traditionally, research into Artificial Intelligence agents for playing games has focused on competitive, perfect information games, such as Checkers~\cite{schaeffer1996chinook}, Chess~\cite{campbell2002deep} and Go~\cite{AlphaGo}. Cooperative games with imperfect information are an interesting research topic not only due to the added challenges posed to researchers, but also because many modern industrial and commercial applications can be seen as examples of cooperation between humans and machines in order to achieve a mutual goal in an uncertain environment. In previous work~\cite{canaan2018towards}, we  surveyed metrics and open problems relating to co-creative and mixed initiative systems, and argued that games, especially cooperative games, are an ideal research platform to address some of these issues.

\textit{Agent modeling} is one of the main features of co-creative and mixed initiative systems identified in this survey, and cooperative games face, to a larger extent than competitive games, the problem of modelling other agents (human players or another AI agents). Usually, the main challenge in agent modeling is about  \textit{predicting} the future actions of other agents, but in our chosen domain it is also important to \textit{interpret} observed actions and infer what they might imply about hidden features of the world. In essence, our agents should be able to represent the mental state of other actors and see the world from their perspective. This ability has also been referred to as having a theory of mind~\cite{premack1978does}.

In this context, interacting with agents for which a model is known in advance is a very different problem than interacting agents for which no such model is known. When playing with known agents, a number of assumptions, or conventions, can be taken for granted, but when playing with \textit{Ad-Hoc teammates}~\cite{stone2010ad,albrecht2018autonomous,foerster2018bayesian}, we need either adaptive strategies that learn and leverage a model of the other players on the fly or non-adaptive strategies that play well with a wide range of partners, despite using the same policy for all of them.

One issue that emerges when dealing with \textit{Ad-Hoc cooperation} scenarios is how to evaluate agents that play in such a setting. A typical approach is to specify a \textit{pool} of agents that we want to be able to play well with. However, if this pool is known in advance, then \textit{challenger} agents can be over specialized towards this specific pool, leading to behavior that might not generalize to other teammates. This typically requires the pool to be kept secret, which can lead to issues of reproducibility. 

Another alternative, which we favor, is to use some stochastic method to generate a pool where agents display enough variety that different strategies are needed to play well with all of them. It is also desirable that the method produces agents with enough variety each time it is run, so it can be reused for multiple experiments without relying on secrecy and without making it too easy to over specialize agents towards it.

In this paper we discuss characteristics that would be desirable in a method for generating pools of agents that can be used to evaluate other agents in ad-hoc cooperation settings. We propose metrics that can be used to characterize these generators and implement a generator of agents using MAP-Elites~\cite{mouret2015illuminating}, a Quality Diversity algorithm that optimizes towards a set of behaviorally diverse, high quality individuals, using a rule-based representation of Hanabi agents. We hope this method will help us better evaluate the challenger agents we want to develop in the future and help others to better evaluate their own agents.


\section{RELATED WORK}

\subsection{Hanabi: the game}

Hanabi is a cooperative card game designed by Antoine Bauza and has won the prestigious \textit{Spiel des Jahres} award for tabletop games in 2013. It is played by groups of 2-5 players who try to play stacks of cards in correct order of rank or value (from 1 to 5) for each of the five colors in the game (B, R, Y, W and G). Players play with the contents of their hands facing outwards, so that each player sees the cards every other player has, but not their own cards. The group can only communicate through information (or hint) actions, which allow the current player to select another player and point to all cards of a chosen rank or color in their hand, at the expense of an information token from a shared pool. When a card is played to the tableau, the group scores a point if it is the next card in its color stack, or loses a life otherwise. A player can also discard a card from their hand, which recovers one information token.


The game ends with a victory for the group if all five stacks are complete with cards ranked 1 to 5 of that color. Whenever a card is played or discarded, players must draw back to their hand limit from a draw deck. If the draw deck is exhausted, every player gets one last turn to take an action, after which the game ends in defeat if not all stacks are complete. The game also ends with defeat if the group loses three lives by playing three invalid cards. The score of the game is the number of cards successfully played by the group, that is, 25 in case of victory or from 0 to 24 in case of defeat.


Hints provide grounded information by disclosing the rank of color of cards. Examples of hints are \enquote{your first, second and fourth card are 1's}, \enquote{your middle card is a 5} and \enquote{your two rightmost cards are yellow}. But in addition to this grounded layer of meaning, players try to glimpse additional implicit meaning from each hint by taking into account a model of the other player. For example, if the  1's of some (but not all) colors have been played, most players who receive a hint of~\enquote{your leftmost card is a two} would assume that that it refers to a playable card, even though nothing was said of its color. The player giving the hint should therefore \textit{predict} how the receiving player would act in different scenarios and the receiving player must in turn \textit{interpret} what each hint (and other actions) says about the state of the game.

Over time, conventions can either emerge organically or be formally agreed to in a group. Conventions provide a guideline of how hints ought to be interpreted and which hints should be given in each situation. Examples of conventions are \enquote{hints should, if possible, identify playable cards rather than unplayable ones} and \enquote{players should discard unidentified cards from oldest to newest}. But these can easily backfire if all players are not on the same page. For AI research, this makes the problem of self-play, where all agents are known to be following the same strategies and conventions, fundamentally different from the problem of ad-hoc cooperation.


\subsection{Hanabi-Playing agents and the competition}

Many of the early approaches~\cite{osawa2015solving,van2016aspects,eger2017intentional} for playing Hanabi with AI were variations of a simple strategy for self-play which prioritizes playing cards that are believed to be playable, followed by giving hints that identify playable cards in other player's hands, followed by discarding cards that are believed to not be necessary. 

Walton-Rivers \textit{et al.}~\cite{walton2017evaluating} are the first to address the problem of  playing with a diverse population of agents with different strategies in Hanabi. First, they reimplemented many previously published agents under a rule-based paradigm, where agents are defined by an ordered sequence of human-crafted rules. Each rule takes a game state and, if a certain condition is true, returns an action. A rule-based agent simply evaluates each rule in order, until it finds a rule that is applicable and executes the corresponding action. They then created a fixed pool of baseline agents using some of these reimplementations, plus some new agents following similar strategies, an agent that takes actions at random, and a \enquote{flawed} that has risky Play actions and gives random hints. Finally, they evaluated each of these agents, plus some tree search agents, based on how well they score on average when successively paired with all agents from the fixed pool.

The 2019 Hanabi CoG competition is based on this work. It first took place at the 2018 CIG conference, where participants submitted agents both for a self-play (or \enquote{mirror}) track and a mixed play track. In the mixed play track, the agents had to play with a pool similar to the one used in~\cite{walton2017evaluating}, but the exact agents were not made public before the competition. The winner of the competition was a variant of Monte Carlo Tree Search~\cite{browne2012survey} by Goodman~\cite{goodman2019re} designed to deal with problems of strategy fusion and nonlocality that arise from executing tree search in a hidden information environment. They also use neural networks to model a pool of other sample agents, and bayesian updates keep track of which agent in this pool best approximates the current partner. It achieved a score of 13.28 in the mixed track and 20.57 in the mirror track~\cite{competition}.

Our 2018 competition entry, which took second place, is described in~\cite{canaan2018evolving}. We implemented an evolutionary algorithm to make rule-based agents by searching for a well-performing sequence of rules both for self-play and mixed play, using the same pool as ~\cite{walton2017evaluating} for mixed play. It achieved a score of 12.85 in the mixed track and 17.52 in the mirror track. While the current paper is based on that previous work, here we ignore the standard pool used in~\cite{walton2017evaluating} and~\cite{canaan2018evolving} and procedurally create our own pool, whose agents are evaluated by how well they fare on self-play and when paired with each other. 

The 2019 competition also features a mirror and mixed track, plus  a new learning track where agents can adapt to repeated play with the same partners. While this paper does not aim directly to develop agents to compete, we hope the pools of ad-hoc partners we are generating can help in evaluating future agents, and in better understanding what types of play work better with agents exhibiting a variety of behaviors.

Other than the CoG competition and its related agents, another work on ad-hoc teamplay using Hanabi is by Bard \textit{et al.}~\cite{bard2019hanabi}, who independently trained reinforcement learning agents that scored 20 to 22 points in self-play, but only 0 to 5 when paired with one another. They also proposed an ad-hoc setting where self-play playtraces of the partner agent are provided prior to gameplay for learning, but no agent currently takes advantage of this feature.

While this paper focuses on ad-hoc play, the best-performing self-play agents we are aware of are BAD by Foerster \textit{et al.}~\cite{foerster2018bayesian} and WTFWThat by Wu~\cite{Wu}. BAD, or Bayesian Action Decoder, uses reinforcement learning to learn a deterministic partial policy shared by two agents, and achieves a score of 23.9 (out of 25) with 2 players. WTFWThat is based on a sophisticated  convention called \textit{hat-guessing} first proposed by Cox \textit{et al.}~\cite{cox2015make} and expanded on by Bouzy~\cite{bouzy2017playing} to achieve self-play scores above 24 with 3 or more players.

\subsection{Quality Diversity and MAP-Elites}

Quality Diversity~\cite{pugh2016quality} (QD) algorithms are a class of population-based search algorithms that aim to generate a large number of behaviorally diverse, high-quality solutions. Diversity of behavior can be pursued either as desirable target in its own right or as an intermediate step to high-quality solutions, as showcased by novelty search, which can outperform traditional optimization in deceptive environments even though it abandons the notion of an objective function~\cite{lehman2011abandoning}. QD differs from novelty search, however, because it does not optimize for novelty alone, but searches for both behavioral diversity and high fitness at once. QD also differs from Multi-Objective Optimization~\cite{deb2014multi}, which searches for trade-offs between one or more objectives, because QD actively attempts to find high-quality solutions in all regions of the behavior space, not just those with good trade-offs.

MAP-Elites~\cite{mouret2015illuminating} is an example of QD algorithm that attempts to \enquote{illuminate} the behavior space by mapping each individual to behavioral \enquote{niche}, while maintaining an archive of the best individual (an elite) in each niche. MAP-Elites was first proposed to pre-compute a variety of effective gaits for a six-legged robot, so that, when the robot suffers damage, it can quickly search for a gait that adapts to the damage and allows it to keep moving at a decent pace~\cite{cully2015robots}.

QD algorithms are a promising approach to our goal of generating pools of partner agents to use in ad-hoc cooperation evaluation for at least three reasons: first, by actively searching for variety, we minimize the risk of coming up with a population of extremely similar agents, that a single narrow strategy can successfully play with. Second, we hope to better model human gameplay, which we expect to exhibit large behavioral diversity. Lastly, by still maintaining a notion of quality, we minimize the risk of coming up with a population of diverse agents which are still very bad at the game, and thus not interesting to play with. 

\section{Definitions and metrics}

\subsection{Definitions}
 
\textit{Ad-hoc Cooperation} is the problem of cooperating with arbitrary, previously unknown \textit{ad-hoc teammates} (human or AI agents). While the general problem of ad-hoc cooperation has been addressed at least since~\cite{stone2010ad}, for the rest of this paper we will focus on ad-hoc cooperation applied to Hanabi. For a more recent survey on the topic, see~\cite{albrecht2018autonomous}.

Naturally, performance in an ad-hoc cooperation setting cannot be properly measured without having a set of ad-hoc teammates in mind. Performance with an agent that takes random actions will be different than performance with a rule-based agent employing very simple heuristics, which will differ from performance with a neural network controller that has implicitly learned very sophisticated conventions.

To differentiate between the agent whose performance is being measured in an experiment, and the agents it is being paired with to measure performance, we define a \textit{pool} of agents to be the set of agents that a \textit{challenger} agent needs to play well with. We  assume that the exact agents in the  pool are not known in advance by the designers of the challenger agent. We define \textit{Ad-hoc performance} to be the challenger's average score when paired with all agents in a pool. All the experiments in this paper were done in a two player setting, either pairing two different agents or two copies of the same agent.

Finally, we define a \textit{generator} to be an algorithm that can output a pool of agents. A stochastic generator might produce a different pool every time it is invoked, so we will call a \textit{run} the act of generating a pool from the generator. Ultimately, being able to produce meaningfully different pools from multiple runs on each invocation is what we expect to make a generator reusable for many  different experiments without relying on secrecy of the underlying algorithm.

\subsection{Metrics}

The metrics below can be applied to individuals in a pool and their generators.

The first metric we are interested in is the \textbf{self-play performance} of the individuals in our pools. While we are ultimately more interested in ad-hoc cooperation than self-play, we believe that self-play scores can give a rough estimation of the overall quality of an agent. Although counterexamples can be constructed (imagine a \enquote{timid} agent that never gives a hint until another player \enquote{breaks the ice} by giving a hint), we expect that in most cases, agents that score very low in self-play will also score low in ad-hoc settings. Agents that score very highly on self-play, however, could have either high or low ad-hoc performance, depending on whether their strategies make strong assumptions about the behavior of other players. 

While we do not prescribe an ideal self-play score range for agents in a pool, we believe the self-play performance is a factor in deciding what kind of experiment one wants to run, and what kinds of adaptations a challenger is expected to accomplish. A pool where fairly simple heuristic agents are mixed with intentionally  bad agents as seen in~\cite{walton2017evaluating} might encourage challengers to identify the bad agents and guard against their bad behaviors, while avoiding strategies that require many assumptions when paired with the regular heuristic agents. On the other hand, a pool of very high-performing self-play agents with sophisticated conventions (such as the ones developed in~\cite{cox2015make,bouzy2017playing} or learned by Reinforcement Learning in~\cite{foerster2018bayesian,bard2019hanabi}) might resemble more a puzzle-solving challenge where the exact convention in use must be identified  to achieve good performance.

Consider a generator that generates a pool of agents with high self-play performance, but where agents exhibit only minimal variation from each other. This generator would be ill-suited for use as benchmark of ad-hoc cooperation since a single non-adaptive strategy could likely play well with all agents in the pool. We define \textbf{intra-run diversity} metrics as metrics that capture how much agents in a pool generated by a single run of the generator differ from each other. Agents might differ either in their underlying representation, such as chromosomes in an evolutionary algorithm or weights in a neural network, or in their observed behavior, which can be captured by how often agents would take the same action in a given game state or other game-specific behavior metrics.

Imagine now an experiment where a generator outputs a pool of very diverse agents, but repeating the experiment always yields the same agents. In this case, if a challenger agent could identify which agent they are playing with, they could effortlessly choose a pre-computed strategy that is effective with that partner. If we want to avoid this, we need variety not only within agents of a single run, but also between agents of different runs. We call this \textbf{cross-run diversity}. In this paper, we are especially concerned with diversity between agents in the same niche, but different runs of the algorithm. We call these agents \textbf{corresponding agents}.

Note that a static pool of agents could possibly have high intra-run diversity, but will necessarily have no cross-run diversity and has to be kept secret between evaluations. A generator with high cross-run diversity, on the other hand, could potentially be challenging to play with even if the method is open to the public.

Besides variety, given a set of agents we can also measure how well they play when paired with each other, that is, their \textbf{average pairwise performance}. Comparing the numeric value of this metric with self-play performance can lead to interesting insights about the agents involved. For example, two agents that score higher when playing with each other than either agent does on self-play might have found a pair of synergistic strategies. Agents in pool with high self-play performances but low pairwise performances could be using strategies that are very effective if shared by both players, but are not suited for ad-hoc play, as is the case in~\cite{bard2019hanabi}. 

 Although producing a competition entry is not our main goal in this paper, high pairwise performance might indicate good candidates for the mixed track, and could also be later used as the basis for an ensemble (or hyper-heuristic) agent that first characterizes the behavior of another agent, then selects the policy that is known to work best with agents in that niche.

\begin{figure*}
\centering
\includegraphics[scale=0.55]{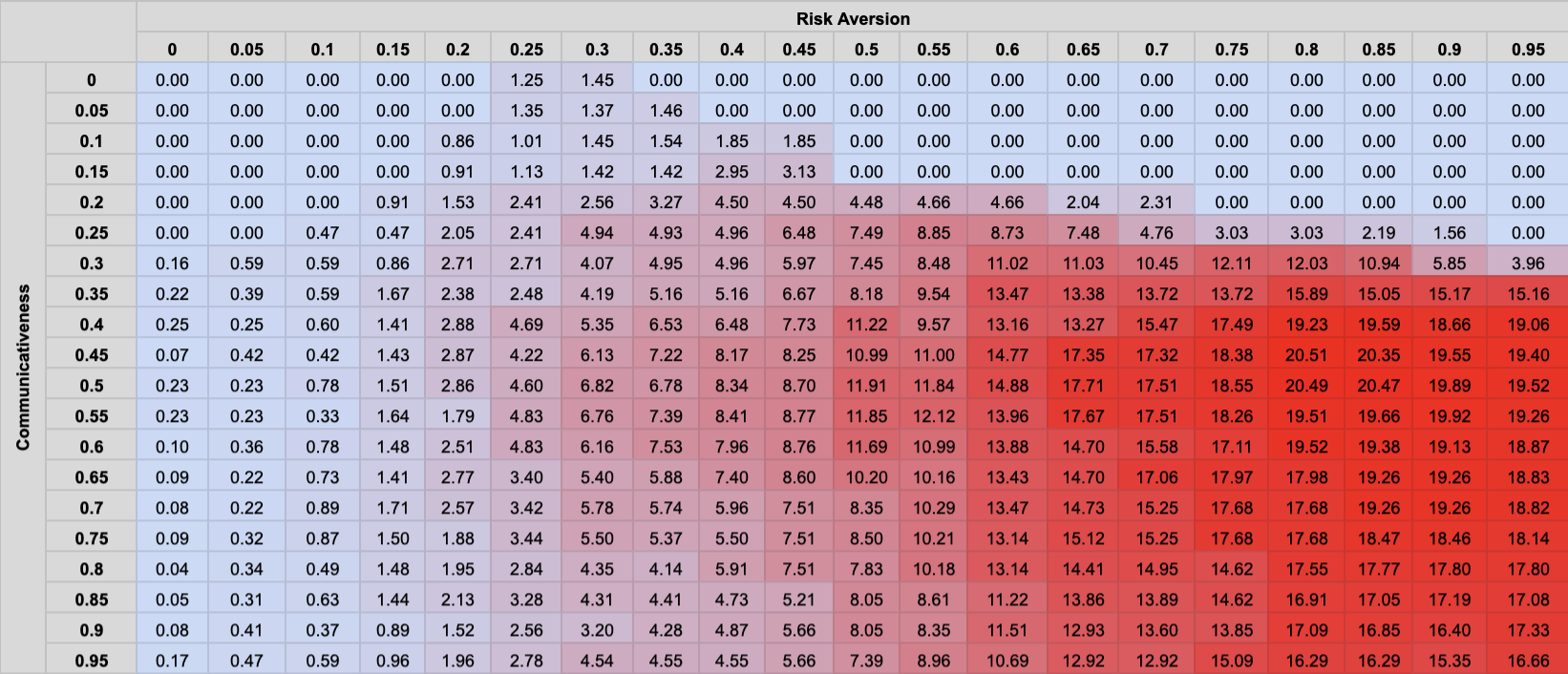}
\caption{Main results of the MAP-Elites experiment after generating 1 million individuals and reevaluating the elite at each niche for 1000 games each. Columns represent risk aversion (most risk averse on the right side) and  rows represent communicativeness (most communicative at the bottom). Values represent the fitness (score) of the best individual in that niche, with redder entries corresponding to higher scores. The maximum score is $20.51$ at  niche (0.45,0.8). The average score is 6.69 across all 400 individuals, or 8.20 when accounting only for the 326 nonzero individuals.}
\label{fig:1Mresults}
\end{figure*}



\section{MAP-Elites Implementation} \label{section:methodology}

Below, we describe how MAP-Elites was implemented to generate populations of diverse Hanabi-playing agents. The implementation will be made public after the competition at our github repository~\footnote{https://github.com/rocanaan/Hanabi-Map-Elites}.

\subsection{Definition of the feature space}

MAP-Elites requires us to select one or more dimensions of variation of behavior that are of interest, which defines a \textit{feature space}. We then need to \textit{discretize} this feature space, which defines \textit{niches} of behavior occupied by agents that exhibit each dimension at a certain range. This allows us to then search for the fittest individual in each niche, also called an elite, as detailed in the next subsection.


We chose the following behavioral metrics as dimensions of the feature space:
\begin{itemize}
    \item \textbf{Risk Aversion:} the average probability that a card is playable, from the perspective of the agent, accounting only for grounded information (and thus no player model), over all game instances where a card was played by the agent. An agent scoring 1 in this dimension would only play cards that are certain to be playable, thus being completely risk averse, while an agent scoring 0 would only purposely play cards that are known to be \textit{un}playable (and so would never score a point through its own play).
    \item \textbf{Communicativeness:} the fraction of time an agent will choose to give a hint, given that it has a hint token available. An agent scoring 1 in this dimension would always give a hint if possible, being fully communicative, while an agent scoring 0 would never give any hints.
\end{itemize}

These dimensions were chosen because they are easy to measure and we believe that they are strategically meaningful, requiring different strategies to play with at different points in the feature space. We also suspected that the highest-scoring behavior in self-play would fall at some value much greater than 0, but lower than 1 for both dimensions: a 0 in either dimension leads to obviously degenerate play, but good play likely requires playing cards under some uncertainty (implying risk aversion $<$ 1) and sometimes passing up the opportunity to give a hint so that another player can better utilize the hint token (communicativeness $<$ 1). 

Note also that while these dimensions help qualify an agent's play, they don't completely determine it. Risk aversion, for example, doesn't tell us \textit{whether} an agent will decide to play a card or do something else, only what's the average probability that a card is playable once it decides to play it. And communicativeness doesn't tell us \textit{which} hint will be given, only the likelihood that \textit{some} hint will be given if a hint token is available.

Each dimension takes values in the range of [0,1], and we chose to discretize them at intervals of 0.05, defining 20 intervals in each dimension, which amounts to 400 niches over all the feature space. And because we are interested in generating agents with varied gameplay, our algorithm will produce the best agent it can find in all the niches, not just the niche that happens to generate the globally optimal agent.

\subsection{Representation and operators}

We use a similar representation of individuals as the one we used in~\cite{canaan2018evolving}. Each individual is represented by a chromosome defined by a sequence of 15 integers, each integer representing one of 135 possible rules. An agent's behavior is determined by simply moving through the rules in the order they appear in the chromosome and selecting the action returned by the first rule that applies. An agent might have rules that never fire during gameplay (for example, a rule that says \enquote{discard a random card} would never fire if it comes after \enquote{discard your oldest card}). An agent can also have duplicate rules, in which case the second instance of the rule will never fire (assuming the rule either fires or not deterministically, which is true for the rules we are using). Nevertheless, these unused or repeated rules are part of an agent's genetic representation and can be passed on to its offspring. We selected 15 as chromosome length because our agents from~\cite{canaan2018evolving} rarely had more than 10 different rules activated.

Mutation is implemented by randomly replacing each rule in a chromosome with a random rule with probability 0.1. Crossover happens with probability 0.5 and is implemented by selecting another individual from the population and randomly selecting (with probability 0.5) the corresponding gene from one of the parents for each position. 

\subsection{Fitness, feature descriptors and niches}

During evolution, each individual is evaluated by playing 100 matches in self-play mode with 2 players, at the end of which we calculate both its fitness and its feature descriptors. Its fitness is the average score achieved in these matches, and its feature descriptor is an ordered pair $(c,r)$, where $0\leq c \leq 1$ is its communicativeness and $0\leq r \leq 1$ is its risk aversion. We then map the individual to a niche, which is a discretization of the behavior space in intervals with size 0.05 in both dimensions. For example, niche (0.4,0.95) has individuals with $0.4 \leq c < 0.45$ and $0.95 \leq r \leq 1$.

Then, we recompute the fitness of the the current elite in the mapped niche by playing another 100 matches in self-play mode. Since evaluation is noisy, we take this extra step to guard against the possibility that the current elite could have had an unusually good run in the previous evaluation. Then a decision is made to either maintain the current elite or the new individual, whichever has the highest fitness for that niche.

\section{Results}



We used Map-Elites to generate and evaluate a total of $10^6$ individuals. The first $10^4$ individuals were initialized with random chromosomes and were meant to initially populate as many of the 400 niches in our map as possible. The remaining individuals were generated by performing mutation and crossover on the current elite of a randomly selected niche. Each individual's fitness and niche were evaluated by simulating 100 matches, as described previously, after which its fitness was compared with the reevaluated fitness of the current elite in the niche. 

While during evolution each individual was evaluated by playing one hundred games, after evolution we reevaluated each elite by playing a thousand self play games. The average score in this reevaluation for each niche's elite is reported in figure~\ref{fig:1Mresults}. On average, reevaluations had a Standard Deviation (SD) of 2.83 (with the highest SD of any individual being 7.21), and Standard Error of the Mean (S.E.M) of about 0.09 with 1000 matches played. 

Our coverage (number of cells filled with an individual scoring above zero) was 326. The average score over the 326 covered niches was 8.20. We report the average over covered niches because only these valid individuals take part in the intra-run and cross-run metrics. For completeness, the average score considering all 400 possible niches would be 6.69. These results are summarized in the first entry of table~\ref{table:runs}. In general, the best agents were found at intermediate levels of communicativeness and high, but not extreme, risk aversion.

The best individual had a score of $20.51$ and was found at niche (0.45,0.8), which corresponds to a communicativeness  between $0.45$ and $0.5$ and a risk tolerance  between $0.8$ and $0.85$. This lends some evidence to our starting assumption that neither behavioral metric is optimal at its highest extreme. This score is also an improvement over our best 2-player agent (called Mirror Situational) reported in~\cite{canaan2018evolving}, which had a reported self-play score of 20.07. We verified this result by doing a new reevaluation of Miror Situational and the (0.45,0.8) elite obtained by MAP-Elites, with 20000 games played each. The results suggest that our new agent is indeed superior, with a reported score in this additional experiment of 20.33 and S.E.M of  0.02 for the (0.45,0.8) elite, versus a score of 20.11 for Mirror situational with the same S.E.M.

Note that 74 niches have a score of zero. This is expected due to the degeneracy of strategy that happens when one of the behavioral dimensions is close to zero. An agent with communicativeness close to zero will almost never give a hint. Agents with very high or very low risk aversion will only play cards that have a high probability of being playable or unplayable, respectively. When paired with a non-communicative agent, they will never have enough information to play a card, justifying the scores of zero in both the top-left and top-right regions of the map.

\subsection{Cross-run evaluations}

One of our main goals is that our method can be used repeatedly to provide pools of agents  for ad-hoc cooperation benchmarking, without relying on secrecy. This requires corresponding individuals to be diverse from run to run, so that it is not easy to predict their actions even if their niche is given alongside agents from previous runs. To verify whether we succeeded, we first performed two new runs of the algorithm, resulting in two new pools of agents. We then paired each individual in each of the three pools with their corresponding individual in the other two pools, calculating what score they achieve when playing together, how similar were their chromosomes and how similar was their behavior. The results of these runs are shown in table~\ref{table:runs}.

\begin{table}[]
\begin{tabular}{|l|l|l|l|l|}
\hline
\textbf{Run} & \textbf{\# Individuals} & \textbf{Max Score} & \textbf{Average Score} & \textbf{Coverage} \\ \hline
Run 1        & 1000000                 & 20.51              & 8.20                   & 326               \\ \hline
Run 2        & 750000                  & 20.68              & 8.27                   & 326               \\ \hline
Run 3        & 147000                  & 20.16              & 7.85                   & 324               \\ \hline
\end{tabular}
\caption{Number of individuals, maximum score, average score (over covered niches) and coverage of each of the three runs.}
\label{table:runs}
\end{table}

We then had each individual from each run play 1000 games paired with the corresponding individual of each other run. The average score of 8.07 we obtained is very close to the average self-play performance of the three runs (8.10), indicating that agents from the same niche, but different runs, play among themselves about as well as they would in self-play.

We then measured the representation diversity and behavior diversity of corresponding agents (in the same niche) across the three runs. To measure representation diversity we computed the Hamming Distance~\cite{hamming1950error} by counting the number of positions at which the genes of corresponding chromosomes differ. Each agent was defined by sequence of 15 ordered rules out of a rule base with 135 rules to choose from. We found the average Hamming Distance to be 14.24 meaning individuals shared less than one rule in the same position on average.

We measured behavior diversity by counting how often each individual takes the same action as their corresponding pair, when presented with the same game state. We first took the 326 valid agents from our first run and recorded all game states found over the course of 100 new games for each individual, for a total of about 140 thousand game states. In these states there was an average of  13 actions to choose from and we observed an action similarity of 0.58.

Our results indicate that corresponding agents play quite well together (comparable to self-play), despite being very diverse both chromosome-wise and taking the same actions as each other less than $60\%$ of the time.


\subsection{Intra-run metrics}

\begin{figure*}
\centering
\includegraphics[scale=0.55]{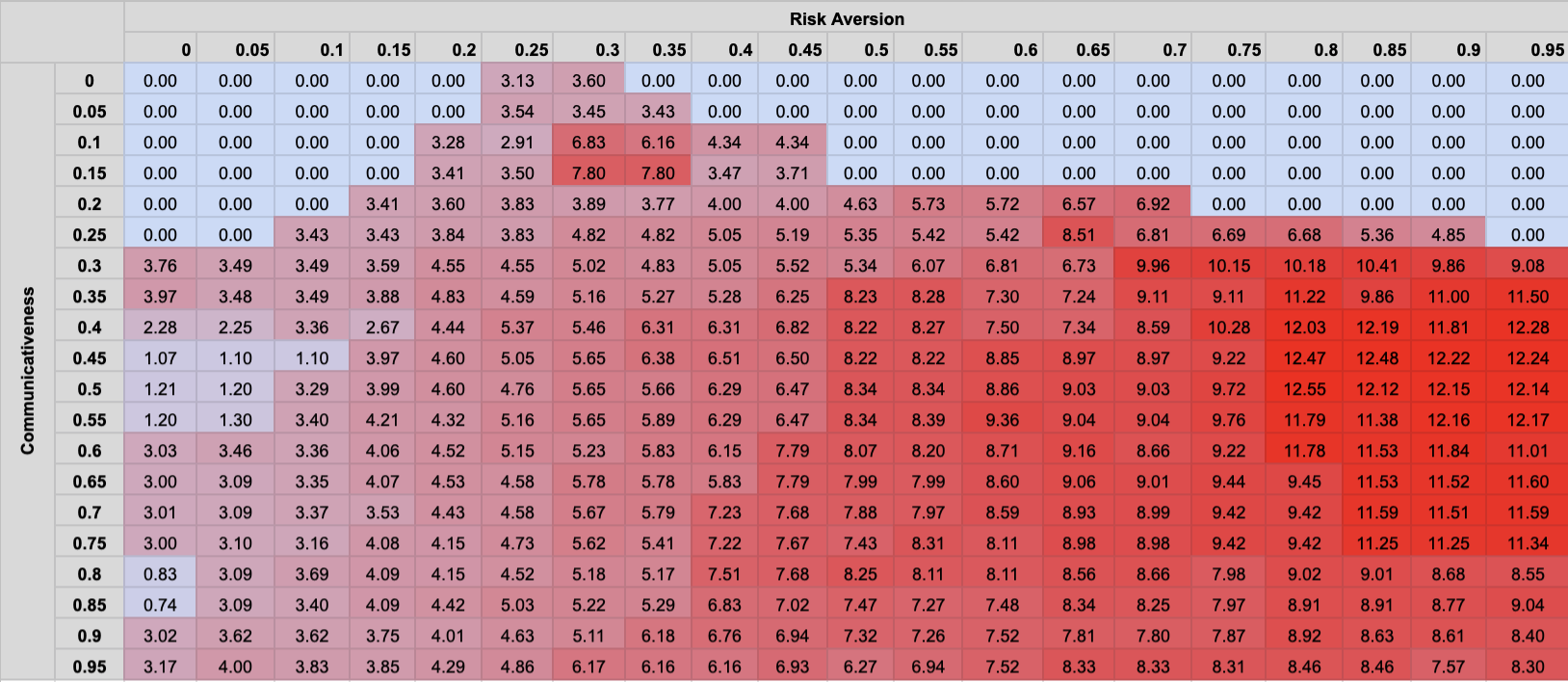}
\caption{Average pairwise performance of each agent when paired with all 326 valid agents in the pool}
\label{fig:IntraScores}
\end{figure*}

\begin{figure*}
\centering
\includegraphics[scale=0.55]{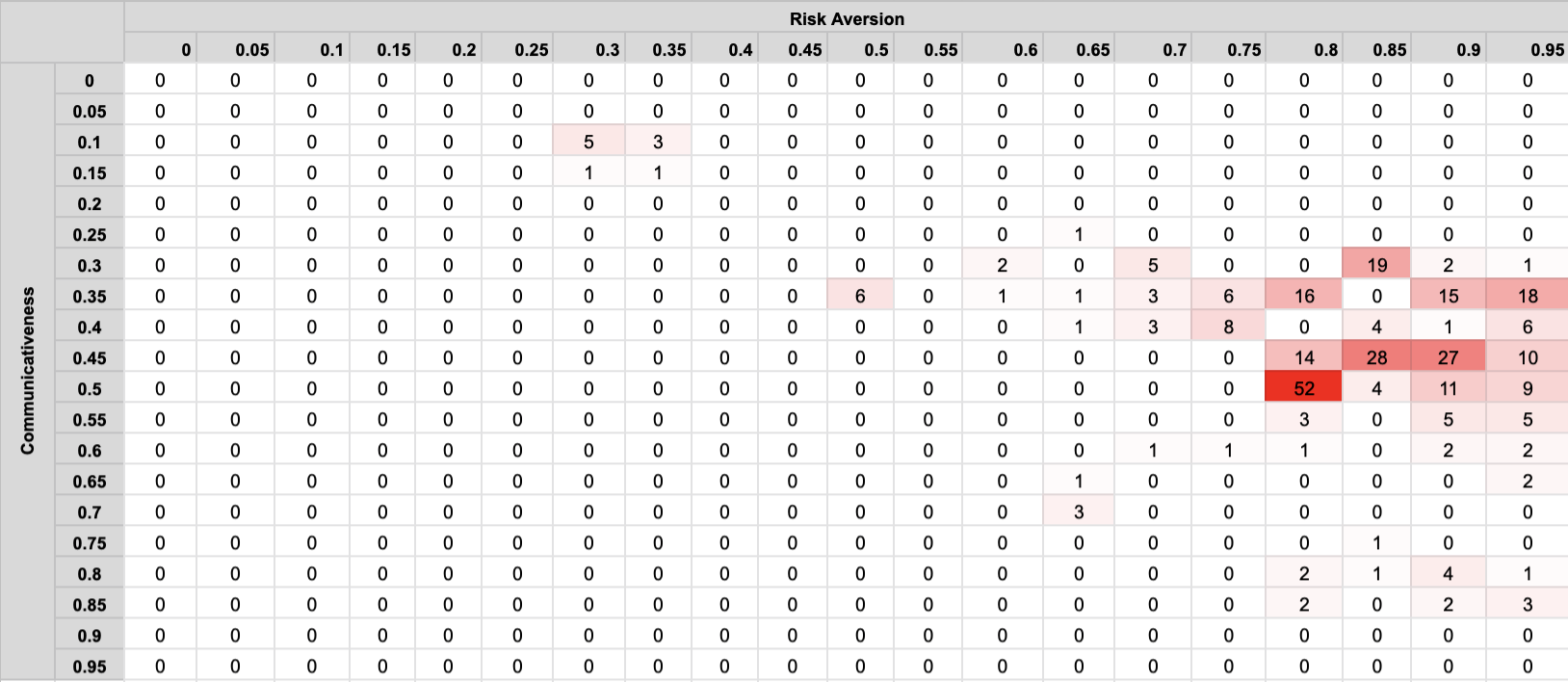}
\caption{Number of pairings for which each agent in the map is the best partner for the other agent.}
\label{fig:BestPartners}
\end{figure*}

Intra-run metrics are important to check if individuals in a single run are varied from each other, and also to generate a map of which behavioral niches play well with each other. Our first step was to evaluate pairwise performance of all $326*326$ pairs of valid individuals. We did so by playing 400 games for each pair, for a total of about 43 million games. We recorded the average score of each agent when paired with all of the pool (including itself). Results are shown in figure~\ref{fig:IntraScores}. We also kept track of which agent was the best partner of each other agent (and possibly itself) in the population, that is, if we fix agent (i,j), which agent (m,n) achieves maximum score with (i.j). The number of times each agent was determined to be the best partner of some agent in the pool is shown in figure~\ref{fig:BestPartners}. 

The results show that, while once again agents with intermediate communicativeness and high, but not extreme risk aversion achieve high scores, agents that were low-performing in self-play achieve higher average pairwise performance than their self-play performance. This suggests that weak agents benefit more from playing with strong agents than from playing with a mirrored strategy. There is also an interesting region of the behavior space, from niches (0.1,0.3) to (0.15, 0.35) that not only achieve better scores than their surrounding neighbors, but were considered best partners for 10 agents. This is specially curious as there doesn't seem to be anything special about this region when looking at the self-play scores. We suspect that agents with very low risk aversion (which play only unplayable cards, but require information to do so) are best paired with low communicativeness agents, as by withholding information, they cannot play the wrong cards.  We verified that all 10 agents that have agents in this region as best partners had risk aversion below 0.2 with 9 of them below 0.1.

\begin{figure}
\centering
\includegraphics[scale=0.5]{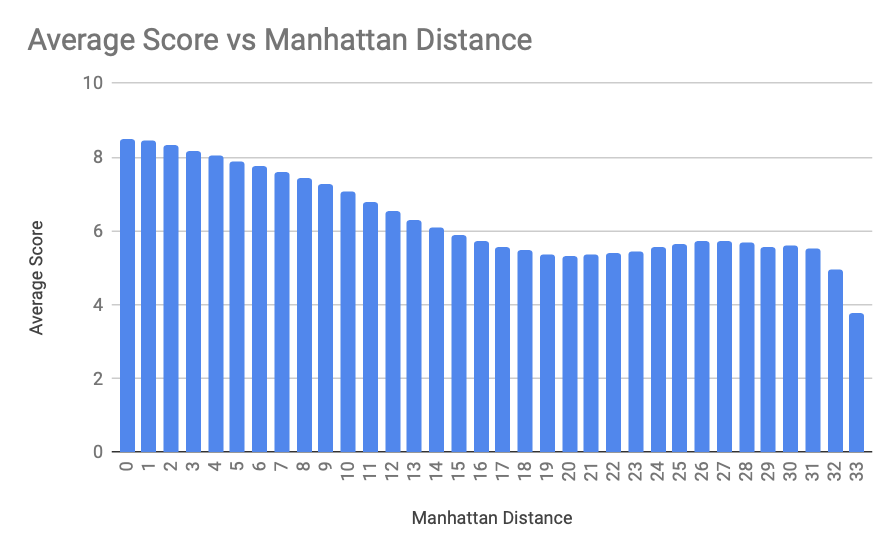}
\caption{Average score of each pair of agent versus the Manhattan Distance of these agents in the behavior space}
\label{fig:IntraHistogram}
\end{figure}

The average pairwise fitness, considering only valid individuals, was 6.64, considerably lower than the average self-play score of 8.20 among valid individuals in the pool. This suggests that, for the most part, agents tend to play better with themselves than with an agent taken randomly from the pool. To verify whether distance in the feature space played a role, we also kept track of how average pairwise performance varied with distance in the feature map. We plotted average pairwise performance against Manhattan Distance in the feature map. The results can be seen in figure~\ref{fig:IntraHistogram}, suggesting that, for the most part, agents play better with themselves or with nearby agents than with faraway agents. Note that no pair of valid individuals had Manhattan Diistance greater than 33.

We also calculated the average intra-run Hamming Distance and Action Similarity. Because of the way we defined the feature space, individuals that are far away from each other will necessarily take different actions and have different chromosomes, but it is still interesting to quantify this diversity. Hamming distance varied from 9.78 between neighboring individuals to 15 for individuals at a Manhattan Distance of 33. Action Similarity varied from 0.66 to 0.07, supporting the assumption that nearby individuals are more genetically and behaviorally similar than faraway individuals.

\section{Future Work}


Our immediate follow-up work is to create challenger agents based on our discretization of the behavior space and the intra-run metrics collected by our experiments. Assuming we can identify the niche of a given individual in a test pool, knowing which agents are likely to make the best partners for each other niche could help us select the most effective policy to play with it. As part of this project, we will take agents previously published in the literature and see which behavior niches they would fit given our methodology, and investigate whether our findings extend to those agents. 

While in this work we measure ad-hoc performance by equally weighting all agents in the pool, in practical applications this might be undesirable either because of the large size of the pool, because some niches have no agents with good enough performance or because the strategies that play well with many might be too similar. The first two issues could be addressed by reducing the number of niches, discarding the niches with low self-play performance or using some method to dynamically adjust the granularity (one of the variations suggested in~\cite{mouret2015illuminating}). The last issue could be addressed by some form of evaluation that is invariant to redundant agents, such as the use off Nash averaging (in the context of 2-player zero-sum games) in~\cite{balduzzi2018re}.

Finally, the current experiments build on a decision-list-like representation, which is easy to evolve but somewhat limited in what policies it can express. We will investigate other evolvable representations, including neural networks and hybrid agents incorporating some tree search.


\section{Conclusion}

We showed that, using MAP-Elites, it is possible to generate a pool of Hanabi-playing agents that differs in two important behavioral dimensions: risk aversion and communicativeness. We could to find well-playing agents within a widely different range of these metrics, but the best-playing agents have moderate communicativeness and high but not extreme risk aversion. The best agent we found in this region had a higher self-play score in 2-player games than the best self-play agent from our 2018 entry. 

The good agents perform best in self-play, whereas bad agents perform best when matched with certain well-playing agents. (Additionally, there is a cluster of agents that are surprisingly collaborative despite low communicativeness, medium-low risk aversion, and bad self-play.) These results hold up across independent replications of the experiment. Importantly, even though individuals in the same cell in different replications of the experiment have different genotypes and play different policies, they play well together, suggesting that the behavioral features identified are relevant measures of playstyle. The results of this paper suggests a way making an ensemble-based (or hyper-heuristic) Hanabi agent which identifies the playstyle of a team-mate and selects an appropriate matching playstyle from its own repertoire.


\section*{ACKNOWLEDGMENT}

Rodrigo Canaan gratefully acknowledges the financial support from Honda Research Institute Europe (HRI-EU).


\bibliographystyle{IEEEtran}
\bibliography{bibfile}
\end{document}